\documentclass[11pt,a4paper]{article}
\usepackage[hyperref]{acl2020}
\usepackage{times}
\usepackage{latexsym}

\usepackage{graphicx}
\usepackage[ruled,vlined,noend]{algorithm2e}
\usepackage{url}
\usepackage{amsmath}
\usepackage{amssymb}

\DeclareMathOperator*{\argmin}{\mathnormal{argmin}} 
\DeclareMathOperator*{\kargmin}{\mathnormal{k-argmin}}
\DeclareMathOperator*{\ngram}{\mathnormal{j-gram}}

\usepackage{microtype}

\aclfinalcopy 

\newcommand*{\affaddr}[1]{#1} 
\newcommand*{\affmark}[1][*]{\textsuperscript{#1}}
\newcommand*{\email}[1]{\texttt{#1}}

\title{Dictionary-based Data Augmentation for Cross-Domain Neural Machine Translation}

\author{%
Wei Peng\affmark[1], Chongxuan Huang\affmark[1], Tianhao Li\affmark[2], Yun Chen\affmark[3], and Qun Liu\affmark[3]\\
\affaddr{\affmark[1]Artificial Intelligence Application Research Center, Huawei Technologies}\\
\email{\{peng.wei1,huang.chongxuan\}@huawei.com}\\
\affaddr{\affmark[2]Beijing University of Posts and Telecommunications}\\
\email{\{litianhao\}@bupt.edu.cn}\\
\affaddr{\affmark[3]Noah's Ark Lab, Huawei Technologies}\\
\email{\{chenyun.nlp,qun.liu\}@huawei.com}%
}

\date{}

\begin{document}
\maketitle
\begin{abstract}
Existing data augmentation approaches for neural machine translation (NMT) have predominantly relied on back-translating in-domain (IND) monolingual corpora. These methods suffer from issues associated with a domain information gap, which leads to translation errors for low frequency and out-of-vocabulary terminology. This paper proposes a dictionary-based data augmentation (DDA) method for cross-domain NMT. DDA synthesizes a domain-specific dictionary with general domain corpora to automatically generate a large-scale pseudo-IND parallel corpus. The generated pseudo-IND data can be used to enhance a general domain trained baseline. The experiments show that the DDA-enhanced NMT models demonstrate consistent significant improvements, outperforming the baseline models by 3.75-11.53 BLEU. The proposed method is also able to further improve the performance of the back-translation based and IND-finetuned NMT models. The improvement is associated with the enhanced domain coverage produced by DDA.
\end{abstract}

\section{Introduction}

Neural machine translation (NMT) models have achieved state-of-the-art translation performance when trained with rich parallel data. A predominant challenge remaining for NMT models is that words in different domains tend to have different meanings \citep{Koehn2017}. Therefore, an NMT model trained in one domain is likely to fail in domains with a significant difference due to domain mismatch \citep{Koehn2017}. Domain adaptation for NMT emerged as critical research to address this issue. A typical practice is to enhance models trained with out-of-domain (OOD) parallel corpora with in-domain (IND) data. As IND parallel corpora are scarce, a category of data-centric domain adaptation research for NMT came into shape  \citep{Chu2018} focusing on 1) leveraging monolingual IND data \citep{zhangandzong2016:1,cheng-etal-2016-semi,Currey2017CopiedMD,domhan}; 2) sentence selection from OOD parallel corpora \citep{wang-etal-2017,van-der-wees-etal-2017-dynamic}; 3) synthesizing IND parallel corpora using back-translation  \citep{sennrich2016a,zhangandzong2016:2}. 

More specifically, one way a domain-specific monolingual corpus is synthesized with an OOD parallel corpus is via direct copying. \citet{Currey2017CopiedMD} directly copy the target monolingual data to the source side of the bitext to train an enhanced NMT model. \citet{cheng-etal-2016-semi} reconstruct the monolingual corpora using an autoencoder, in which both source and target monolingual data can be used. The multi-task learning framework is also applied to joint-train the NMT models with IND monolingual data and OOD parallel corpus \citep{zhangandzong2016:1,arcan}.  On the other hand, data augmentation from selecting sentences similar to IND sentences from OOD has been adopted \citep{wang-etal-2017}. Along this line, dynamic data selection \citep{van-der-wees-etal-2017-dynamic} is proposed as a way to change the selected subsets of training data between training epochs resulting in reported improvements over baseline models.   

Another stream of research works generates synthetic parallel corpora by back-translating target-side or/and source-side monolingual data \citep{sennrich2016a,zhangandzong2016:2}. Back-translation (BT) proves an effective data augmentation solution for cross-domain NMT, and it becomes the default method for many current NMT systems.   

In this paper, we address data augmentation for domain adaptation of NMT by leveraging widely available domain-specific dictionaries from the translation industry. This work is motivated by an observation that existing data augmentation research reliant on IND monolingual corpora and back-translation still suffer from issues associated with the domain information gap. More specifically, unlike OOD bitext, IND monolingual corpora are not always available.  In addition, each IND corpus covers a limited scope of terminology. For instance, we find EMEA (European Medicines Agency) corpus contains only around $10\%$ of the medical terms offered by a subset of domain dictionary derived from SNOMED-CT  \footnote{SNOMED-CT is a collection of multilingual clinical healthcare terminology in the world.  https://www.nlm.nih.gov/healthit/snomedct/us\_edition.html}. As a result, the issues caused by rare or out-of-vocabulary (OOV) terminology cannot be handled well by these methods. Incorporating domain-specific dictionaries appears a promising step towards addressing this issue as it will bring new domain information into NMT. We propose a dictionary-based data augmentation (\textbf{DDA}) method for cross-domain NMT to this end. 

Existing lexicon incorporation research works focus on applying lexicons to only replace rare words in the training bitext \citep{zhangandzong2016:2, fadaee-bisazza-monz:2017:Short2}. We argue such methods are likely to limit the benefits offered by dictionary-based data augmentation as there is no additional IND information included. This research proposes an effective and practical way to automatically generate large-scale pseudo-IND parallel corpora by substituting the selected bilingual lexicon/phrase pairs from OOD parallel corpus with all potential terminology from IND dictionaries. The generated pseudo-IND parallel corpus can be used to enhance an OOD trained baseline NMT model. To the best of our knowledge, no data augmentation research based on lexicon incorporation has attempted to leverage IND dictionaries and OOD corpora in this manner.   The proposed \textbf{DDA} approach brings forth the following contributions:

\begin{itemize}
  \item Without relying on IND monolingual corpora, the proposed approach uses easily-accessible OOD corpora and domain dictionaries instead. DDA can automatically generate large-scale pseudo-IND data for low frequency and OOV terminology to improve translation accuracy;
  \item In contrast to the existing lexicon incorporation methods focusing on IND monolingual corpora and BT,  DDA incorporates all potential IND terminology from a dictionary. This leads to enhanced domain information integration during data augmentation. As a result, DDA-enhanced models can translate terminology not covered by BT-based approaches. DDA is a viable complement to BT-based methods in data augmentation for cross-domain NMT.
\end{itemize}

The experiments are conducted on four translation directions covering English $\leftrightarrow$ French and English $\leftrightarrow$ German language pairs across a general domain (news) and the medical domain. The DDA-enhanced NMT models demonstrate consistent significant improvements, outperforming the baseline models by \textbf{3.75-11.53} BLEU scores. In addition, when combined with back-translation, the proposed method can further improve the cross-domain translation BLEU scores by up to \textbf{18.59}. Furthermore, DDA is able to significantly improve the BLEU scores of an NMT model finetuned with an IND parallel corpus by a range between \textbf{2.34} and \textbf{8.00}. 

\begin{figure*}[]
    \centering
    \includegraphics[width=0.9\textwidth]{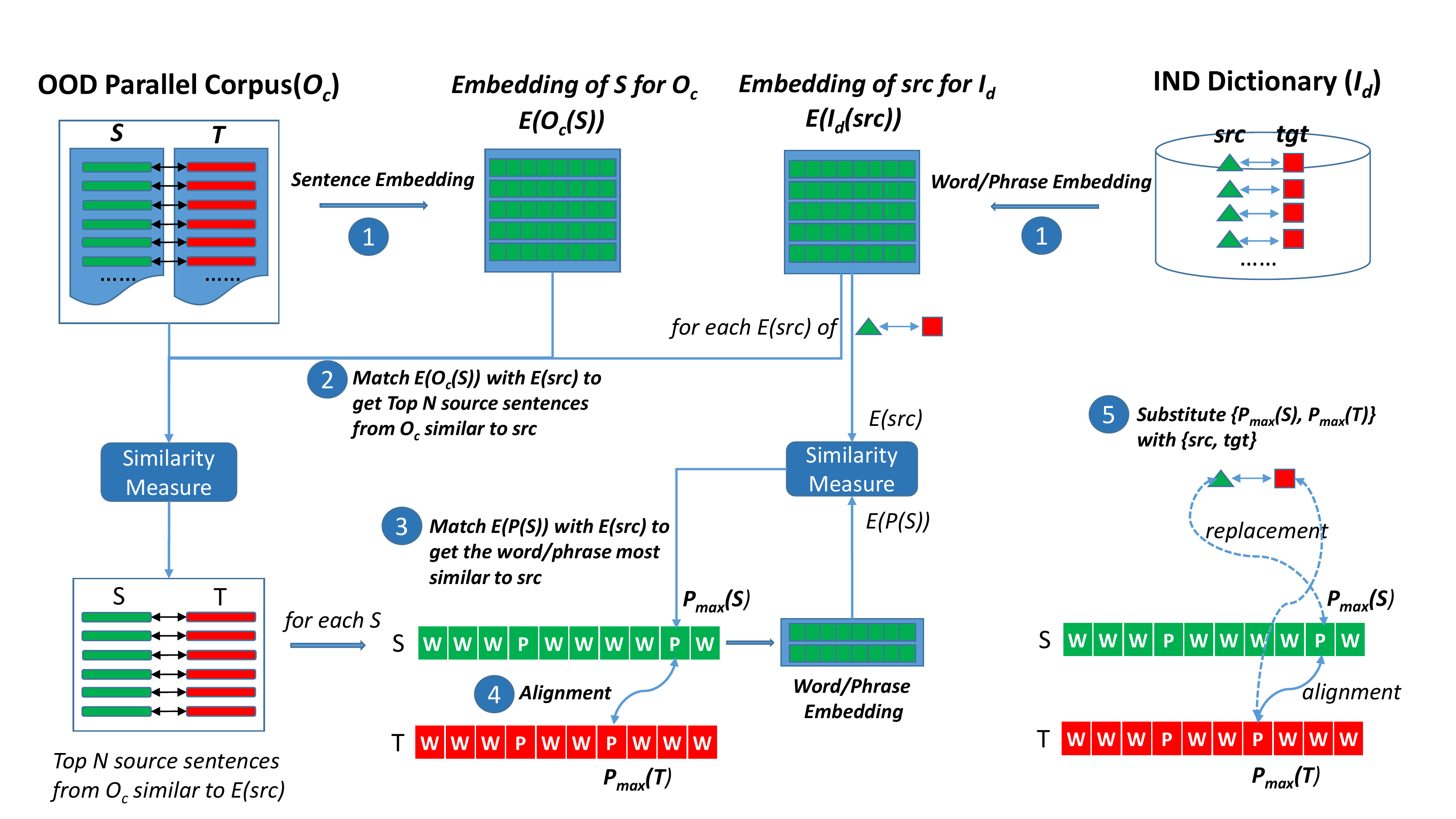}
    \caption{The proposed DDA for cross-domain NMT;  $S$, $T$ refer to source and target sentences for the OOD parallel corpus ($O_c$). $W$ and $P$ stand for a word and a phrase within sentences of the $O_c$. $src$ and $tgt$ depict the source and target words/phrases of the IND Dictionary ($I_d$). $P_{max}(S)$ is the word/phrase with the maximal similarity to the embedding of $src$. 
    }
    \label{figure1}
\end{figure*}

\section{Related Work}
The exploration of integrating bilingual lexicons currently focused on improving rare words for low-resource NMT by applying back-translation to \citep{zhangandzong2016:2} or developing language models on monolingual data \citep{fadaee-bisazza-monz:2017:Short2}. The pseudo sentence pair synthesis method shown in \citet{zhangandzong2016:2} develops a phrase-based machine translation model (PBMT) using the parallel corpus and the dictionary. The developed PBMT is subsequently used to translate sentences from a source monolingual corpus containing the dictionary lexicons. The original bitext and the pseudo sentence pairs are synthesized together to train a new NMT model resulting in improvements in BLEU scores. However, this approach is not designed for cross-domain translation; therefore, it is not in the scope of this study.    

A data augmentation approach for low-resource NMT focuses on generating new sentence pairs for low-frequency words \citep{fadaee-bisazza-monz:2017:Short2}. This study uses a monolingual corpus to train a language model (LM), which subsequently suggests a rare word substitution. The rare word replacement is implemented by a statistic machine translation model and the LM model. This study does not address data augmentation for cross-domain NMT. Furthermore, the baseline model is trained using a largely under-sampled bitext to simulate low-resource NMT scenarios, resulting in sub-optimal BLEU scores (i.e., below 15.4 for DE2EN) which is unfeasible to be deployed in real cross-domain NMT practice.

Semi-supervised lexicon induction is also applied to domain adaptation \citep{hu-etal-2019-domain}. This study applies existing supervised \citep{xing-etal-2015-normalized} and unsupervised \citep{lample2018word} lexicon induction methods to learn a lexicon. The lexicon is further used to construct a pseudo-parallel IND corpus via word-for-word back-translating monolingual IND target sentences into source sentences. Despite shown promising results in domain adaptation for NMT, only one translation direction (DE2EN) has been explored. More recently, Human Annotated Bilingual Lexicons (HABLex) has been released and further explored to incorporate bilingual lexicons into NMT systems \citep{thompson-etal-2019-hablex}. A human aligned and annotated bilingual lexicon is directly used to fine-tune an NMT model under the Elastic Weight Consolidation (EWC) training method \citep{Kirkpatrick3521}. The lexicon is directly used for fine-tuning the baseline without integrated with any OOD bitext. There are also studies exploring the integration of IND lexicons via constrained decoding. For example, \citet{hokamp-liu-2017-lexically} uses a domain-specific terminology to generate target-side constraints to enable a general domain model to adapt to a new domain with significant performance gains. Another study focuses on enhancing terminology translation via replacing the out-of-vocabulary (OOV) words with domain-specific lexicons after decoding \citep{arcan}. They are related to but not in the scope of this study.



\section{Dictionary-based Data Augmentation for Cross-domain NMT}
This method focuses on how to use a domain-specific dictionary ($I_d$) and an OOD parallel corpus ($O_c$) to create a pseudo-IND parallel corpus ($G_c$). We train a baseline NMT using an OOD parallel corpus, which is further enhanced by training on a mixture of the generated pseudo-IND parallel corpus and the OOD bitext (aka the mixed bitext). The baseline model is based on the transformer \citep{Vaswani2017} architecture consisting of an encoder stack and a decoder stack with a multi-head attention mechanism. 

The data augmentation method involved in generating a pseudo-IND parallel corpus based on a domain-specific dictionary is depicted in Figure~\ref{figure1} and Algorithm 1. There are a number of steps including phrase and sentence embedding, matching, alignment and substitutions. As shown in Figure~\ref{figure1}, Step 1 depicts the process of calculating word/phrase embedding from $I_d$ and sentence embedding from $O_c$. The phrase-sentence matching process is done by Step 2 to produce top N source sentences similar to $src$. In Step 3, the word/phrase most similar to $src$ is identified in $S$ with the position of its counterpart in $T$ identified by the alignment model in Step 4. Step 5 shows the process of replacement.

\begin{algorithm}[]
\SetAlgoLined
\SetKwInOut{Input}{Input}\SetKwInOut{Output}{Output}
\Input{ \# IND dictionary \\ $I_d=\{(src^i,tgt^i)\}_{i=1}^m$ \\
OOD parallel corpus \\ $O_c=\{(S^j,T^j)\}_{j=1}^n$}
\Output{ \# Pseudo-IND parallel corpus \\ $G_c=\{(G_{src}^k,G_{tgt}^k)\}_{k=1}^o$}
 $G_c \leftarrow \emptyset$ \\
 \# Compute embedding for input data:\\
 $E(I_d(src))=\{E(src^i)\}_{i=1}^m$ \\
 $E(O_c(S))=\{E(S^j)\}_{j=1}^n$ \\
 \For {$E(src) \in E(src^i)_{i=1}^m$} 
 {
  \# $Top N$ sentences similar to $src$:\\
 $(S^k)_{k=1}^N \leftarrow faiss(E(src),E(S^j )_{j=1}^n)$ \\
  \# For each sentence similar to $src$: \\
 \For {$S^k \in (S^k)_{k=1}^N$} 
 {
  \# Word/phrase $P$ for each sentence: \\
 $P(S^k) \leftarrow phrase\_extract(S^k)$ \\
  \# Word/phrase most similar to $src$:\\ 
 $P_{max}(S^k) \leftarrow  top\_sim(E(src),E(P(S^k)))$ \\
  \# Get counterpart in target sentence:\\
 $P_{max}(T^k) \leftarrow align(P_{max}(S^k))$ \\
  \# Finalize replacement:\\
 $G_c(S) \leftarrow sub(S^k, P_{max}(S^k), src) $ \\
 $G_c(T) \leftarrow sub(T^k, P_{max}(T^k), tgt) $ \\
 $G_c = G_c \cup \{(G_c(S),G_c(T))\}$
 }
 }
 \KwRet{$G_c$}
 \caption{Pseudo-IND parallel corpus generation}
\end{algorithm}

\subsection{Phrase and Sentence Embedding}
Firstly, we obtain the embedding of the source phrase from the domain-specific dictionary, denoted as $E(I_d(src))$, of the IND dictionary ($I_d$) and the embedding of the source sentences, $E(O_c(S))$, from the OOD parallel corpus ($O_c$). The bert-as-service \citep{xiao2018bertservice} with the pre-trained multilingual cased BERT-base \citep{devlin2018bert} has been used as service to calculate the embedding for phrases and sentences from the IND dictionary and the OOD parallel corpus. The BERT-base version of the multilingual cased model is adopted. \footnote{Model details: a 12-layer encoder stack, 768-hidden, 12-heads, 110M parameters. Trained on cased text in the top 104 languages with the largest Wikipedias.} 
It is noted that only the source word/phrase of the dictionary ($src$) and the source sentence ($S$) from the OOD bitext are used for computation in this study. The target word/phrase ($tgt$) may also be applied to generate a likely slightly different pseudo-IND parallel corpus.    

\subsection{Phrase-sentence Matching}
The purpose of phrase-sentence matching is to select sentences from the OOD parallel corpus to serve as the host template for us to implant domain terminology from the IND dictionary. For the embedding of each source phrase of the IND dictionary ($E(src)$), we search for the $top N$ OOD source parallel sentences (${(S^{k})}_{k=1}^N$) based on their similarity to IND source phrase embedding, as shown in Equation~\eqref{eq:cos}. $Faiss$ \citep{JDH17} is adopted to perform efficient similarity search due to a large volume of data involved. $Faiss$ is a library for similarity search in large datasets via developing indexes over the clustering of dense vectors. Cosine similarity (inner product) is applied to both the clustering and search process. The above step produces a list of template sentences ${(S^{k})}_{k=1}^N$ for further processing.

\begin{eqnarray}
\label{eq:cos}
(S^k)_{k=1}^N = \kargmin_{j} \frac{E(src)\cdot E(S^j)}{\left \| E(src) \right \|\cdot \left \| E(S^j) \right \|}
\end{eqnarray}

\subsection{Phrase Matching, Alignment and Substitution}
After tokenization, the phrases for each host template sentence are extracted using TextBlob \footnote{https://github.com/sloria/TextBlob/} with their embedding compared to that of the IND source phrase. The $top\_sim$ function shown in Algorithm 1 extracts the word/phrase with the maximal cosine similarity to the matched terminology from the IND dictionary (Equation~\eqref{eq:cos2}). We focus on nouns in this study because a majority of terms from the IND dictionary are noun words/phrases. 

\begin{eqnarray}
\label{eq:cos2}
P_{max}(S^k) = \argmin_{P} \frac{E(src)\cdot E(P(S^k))}{\left \| E(src) \right \|\cdot \left \| E(P(S^k)) \right \|}
\end{eqnarray}

An alignment model \citep{dyer-etal-2013-simple} is trained based on the OOD parallel corpus to locate the words or phrases ($P_{max}(T^k)$) from the target sentences, which correspond to the candidate phrases ($P_{max}(S^k)$) from source host template sentences. By  substituting  the  candidate  word/phrase  pair \{$P_{max}(S^k),P_{max}(T^k)$\} from the host template sentence pairs \{$S^k,T^k$\} with the matched phrase pairs from the IND dictionary \{$src^i,tgt^i$\}, pseudo-IND parallel sentence pairs can be generated. Iterating all dictionary entries, Algorithm 1 produces a pseudo-IND parallel corpus.   

\begin{table}[b]
\begin{center}
\begin{tabular}{ll}
\hline \textbf{Hyperparameters} & \textbf{Values} \\ \hline
Encoder Layers & 6 \\
Decoder Layers & 6 \\
Embedding Units & 2,048 \\
Attention Heads & 8 \\
Feed-forward Hidden Units & 512 \\
Initial Learning Rate & 0.0007 \\
Train Steps & 100,000 \\
Vocab EN/FR & 43,244 \\
Vocab EN/DE & 43,756 \\
\hline
\end{tabular}
\end{center}
\caption{\label{para-table} Hyperparameters for this study. }
\end{table}

\begin{table*}
\centering
\begin{tabular}{lcc}
\hline \textbf{Dataset} & \textbf{Corpus} & \textbf{Number of Lines} \\ \hline
Train Dataset (OOD) & \begin{tabular}[c]{@{}c@{}}Europarl,News-commentary, \\ Common Crawl\end{tabular} & \begin{tabular}[c]{@{}c@{}}5,394,261 (EN/FR)\\ 4,475,414 (EN/DE)\end{tabular} \\
Dev Dataset (IND) & EMEA & 5,000 \\
Test Dataset (IND) & EMEA & 1,000 \\
\hline
\end{tabular}
\caption{\label{data-table} Datasets used for training and testing the baseline model. }
\end{table*}

\section{Experiment Settings}
The experiments are performed on four translation directions for English $\leftrightarrow$ French (EN/FR) and English $\leftrightarrow$ German (EN/DE) language pairs. The OOD data used for pre-training for the baseline model are extracted from WMT 14\footnote{http://www.statmt.org/wmt14/translation-task.html} including Europarl V7, New-commentary V9 and Common Crawl corpora. The baseline model is evaluated on random IND samples from EMEA, which contains medical documents from the European Medicines Agency. These data are obtained from Opus  (Tiedemann, 2012) and WMT 19 biomedical shared task website \footnote{http://www.statmt.org/wmt19/biomedical-translation-task.html}. Table~\ref{data-table} depicts the data involved.

The IND dictionary is prepared from SNOMED-CT medical concepts with a size of 36,809 (EN/DE) and 36,306 (EN/FR), respectively.  

The generated pseudo-IND parallel corpora for EN/DE and EN/FR are of various sizes to explore their effects on cross-domain NMT.  These generated data along with the OOD parallel corpus are cleaned and pre-processed using functions from Moses \citep{moses}. The punctuation is normalized into standard forms. The tokenization function breaks down sentences into processing units, which are tokens in this study. True-case models are trained to adjust the casing of the initial words for each sentence. Byte pair encoding (BPE) \citep{sennrich2016b} is applied to deal with out-of-vocabulary rare words. The experiment is based on the transformer architecture \citep{Vaswani2017} in fairseq \citep{ott2019fairseq} \footnote{https://github.com/pytorch/fairseq/}.  The hyperparameters are captured in Table~\ref{para-table}. The generated pseudo-IND parallel corpus is used to be mixed with the OOD bitext (aka the mixed bitext) to train enhanced models. 

To compare with a related work using lexicon induction for domain adaptation in NMT, we adopted pseudo-IND parallel corpus generation method recently presented in DALI \citep{hu-etal-2019-domain}\footnote{https://github.com/junjiehu/dali}. This method learns lexicons from EMEA training data and constructs a phrase-based translation model to back-translate monolingual target sentences from EMEA. The DALI NMT model is trained with mixed data from the OOD bitext and DALI-generated pseudo-IND parallel corpus (with 1 million lines in this study). 

Another experiment is performed to demonstrate the complementary effect of the proposed approach to methods using back-translation (BT), in which the target sentences of EMEA are back-translated using the baseline model with their BLEU scores reviewed. It is noted the EMEA data is cleaned to remove duplication resulting in corpora with a size of 287,738 (EN/DE) and 284,536 (EN/FR), respectively. The experiment results (in Appendix~\ref{sec:appendix}) confirm that deduplicating EMEA leads to higher BLEU scores.  

The results obtained from fine-tuning the baseline models with the IND bitext (deduplicated EMEA) are defined as IND-finetuned (IND-FT) results. We also examine the effect of DDA on the IND-FT NMT models. This is performed by training the NMT model with the mixed data of OOD bitext and the DDA-generated data (DDA-G) and subsequently fine-tuning the model with the IND bitext EMEA.    

The BLEU scores are measured using the score.py from \textit{fairseq}, which implements sacrebleu.py \citep{post-2018-call}.

\section{Results and Analysis}

\begin{table*}[]
\centering
\begin{tabular}{lcccc}
\hline
\multicolumn{5}{c}{\textbf{BLEU on EMEA: Effects on Baseline}} \\ 
\hline
\textbf{Models} & \textbf{EN2FR} & \textbf{FR2EN} & \textbf{EN2DE} & \textbf{DE2EN} \\ \hline
Baseline & 25.73 & 24.91 & 23.98 & 29.41 \\ 
Baseline + DALI-G & - & - & - & 33.98 (+4.57) \\ 
Baseline + BT & 33.14 (+7.41) & 27.76 (+2.85) & 31.53 (+7.55) & 36.93 (+7.52) \\ 
Baseline + DDA-G & 30.92 (+5.19) & 36.44 (+11.53)  & 28.24 (+4.26) & 33.16 (+3.75) \\ 
Baseline + BT + DDA-G & 37.96 (+12.23) & 43.50 (+18.59) & 35.04 (+11.06) & 40.60 (+11.19) \\
\hline
\multicolumn{5}{c}{\textbf{Effects on IND-finetuned (IND-FT) NMT}} \\ 
\hline
IND-FT: Baseline + EMEA & 45.90 & 53.38 & 48.71 & 55.68 \\ 
IND-FT + DDA-G & 53.90 (\textbf{+8.00}) & 58.22 (\textbf{+4.84}) & 52.13 (\textbf{+3.42}) & 58.02 (\textbf{+2.34}) \\ 
IND-FT + BT + DDA-G & 54.12 (\textbf{+8.22}) & 58.64 (\textbf{+5.26}) & 53.25 (\textbf{+4.54}) & 58.63 (\textbf{+2.95}) \\ 
\hline
\end{tabular}
\caption{\label{result-table} Experimental results for this study. DALI-G depicts that the pseudo-IND parallel corpus (1 million lines) are generated using DALI. DALI focused only on the DE2EN translation direction. DDA-G describes the pseudo-IND parallel corpus (1 million lines) are generated by the proposed method. BT stands for back-translation using the target sentences from the deduplicated EMEA training data. The \textbf{bold numbers} show the effects of DDA on the IND-finetuned (IND-FT) NMT models. }
\end{table*}

Table~\ref{result-table} gathers the results of a range of experiments mentioned above. The baseline models trained using the OOD bitext (news) are evaluated against IND medical data EMEA. The proposed DDA approach is used to generate a pseudo-IND parallel corpus (with 1 million lines) to enhance cross-domain NMT performance. This method significantly improves the translation performance of the baseline models with their BLEU scores increased by a range from \textbf{3.75} to \textbf{11.53} (Baseline + DDA-G from Table~\ref{result-table}). 
DDA method can match the results delivered by BT-based methods using IND monolingual corpus EMEA. When combined with BT (Baseline + BT + DDA-G), DDA can further improve the BLEU scores of the baseline models by up to 18.59 (FR2EN), outperforming BT-based methods. DDA is confirmed to be a viable complement to BT-based methods in data augmentation for cross-domain NMT. As for the IND-finetuned (IND-FT) NMT system trained with the OOD bitext and fine-tuned with EMEA, DDA can improve the BLEU scores by a range between 2.34 and 8.00. 
A specific example is shown in Table~\ref{example-table}. The baseline and BT do not produce the correct English translation for the medical term \textcolor{blue}{``\'enalaprilate"}. The correct translation of the term \textcolor{blue}{``enalaprilat"} can be generated by a DDA-enhanced baseline. 

\begin{table*}[h]
\centering
\begin{tabular}{ll}
\hline
\multicolumn{2}{c}{\textbf{FR2EN Source and Reference}} \\ \hline
\textbf{Source} & \begin{tabular}[c]{@{}l@{}}Dans la fourchette de concentrations en rapport avec les doses \\ th\'erapeutiques , la liaison aux prot\'eines plasmatiques de l' \\\textcolor{blue}{\'enalaprilate} ne d\'epasse pas 60 \% .\end{tabular} \\ \hline
\textbf{Reference} & \begin{tabular}[c]{@{}l@{}}Over the range of concentrations which are therapeutically \\ relevant , \textcolor{blue}{enalaprilat}  binding to human plasma proteins does \\not exceed 60 \% .\end{tabular} \\ \hline
\multicolumn{2}{c}{\textbf{Examples of Translation Results}} \\ \hline
\textbf{Baseline} & \begin{tabular}[c]{@{}l@{}}In the range of therapeutic dose - related concentrations , the \\ link to plasma proteins in \textcolor{red}{enalaprilate} does not exceed 60 \% .\end{tabular} \\ \hline
\textbf{Baseline + BT (EMEA)} & \begin{tabular}[c]{@{}l@{}}In the therapeutic dose concentration range , binding to plasma \\proteins of the nearest $<$unk$>$; generic \textcolor{red}{enalaprilate} does not \\ exceed 60   \% .\end{tabular} \\ \hline
\textbf{Baseline + DDA-G} & \begin{tabular}[c]{@{}l@{}}In the range of concentrations related to the therapeutic doses , \\ the link to the plasma proteins of \textcolor{blue}{enalaprilat} does not exceed \\ 60 \% .\end{tabular} \\ \hline
\textbf{IND-FT + BT (EMEA) + DDA-G} & \begin{tabular}[c]{@{}l@{}} In the therapeutic dose - range , binding of \textcolor{blue}{enalaprilat} to plasma\\ proteins does not exceed 60 \% .\end{tabular} \\ \hline
\end{tabular}
\caption{\label{example-table} An example of translation output for EMEA demonstrates the effect of DDA on FR2EN NMT. The fonts in blue indicate the ground truth of the medical term and the red fonts show incorrect translation. }
\end{table*}

\section{Further Analysis}
The effect of the size of the IND dictionary is further studied in this section. The experiment is performed by comparing the BLEU scores of enhanced models trained with the mixed bitext generated by IND dictionaries of various numbers of entries (i.e., pairs). Apart from BLEU, we calculate enhanced domain coverage to quantify the effect of incorporating an IND dictionary on baseline models. The enhanced domain coverage ($ED$) of a dataset ($C$) for an IND dictionary ($I_d$) is defined as the number of unique terms  (1-5 gram starting from a word $w$) captured in the dictionary, which also appear in the dataset $C$ and the test dataset ($Test$):

\begin{figure}[h]
    \centering
    \includegraphics[width=0.47\textwidth]{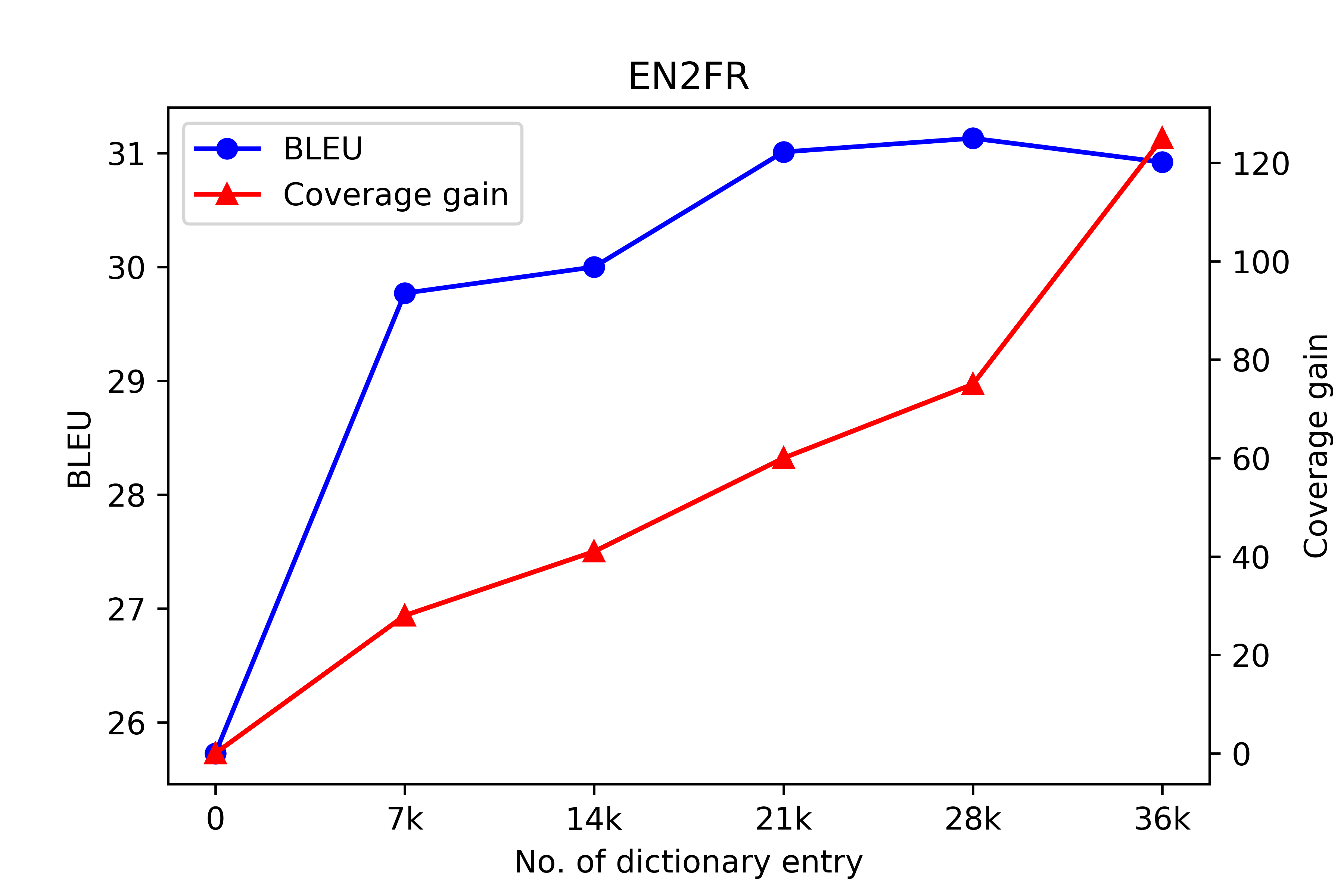}
    \caption{The effects of DDA under various sizes of IND dictionaries for EN2FR models. The second y-axis, `Coverage gain' indicates the gain of enhanced domain coverage of the DDA-generated pseudo-IND corpus with 1 million lines  $ED((DDA-G) \cap Test)$ over that of the OOD training data $ED(Train \cap Test)$ with around 5.4 million lines.}
    \label{figure2}
\end{figure}

\begin{eqnarray}
\label{eq:dc}
ED(C \cap Test) = \sum_{i=1}^{n} \sum_{j=1}^{5} \ngram (w_{i}) 
\end{eqnarray}
\noindent where the available 1-5 gram $\ngram (w_i) \in (I_d \cap C \cap Test)$, $n$ is the number of terms fulfilling the condition.

\begin{figure}[h]
    \centering
    \includegraphics[width=0.47\textwidth]{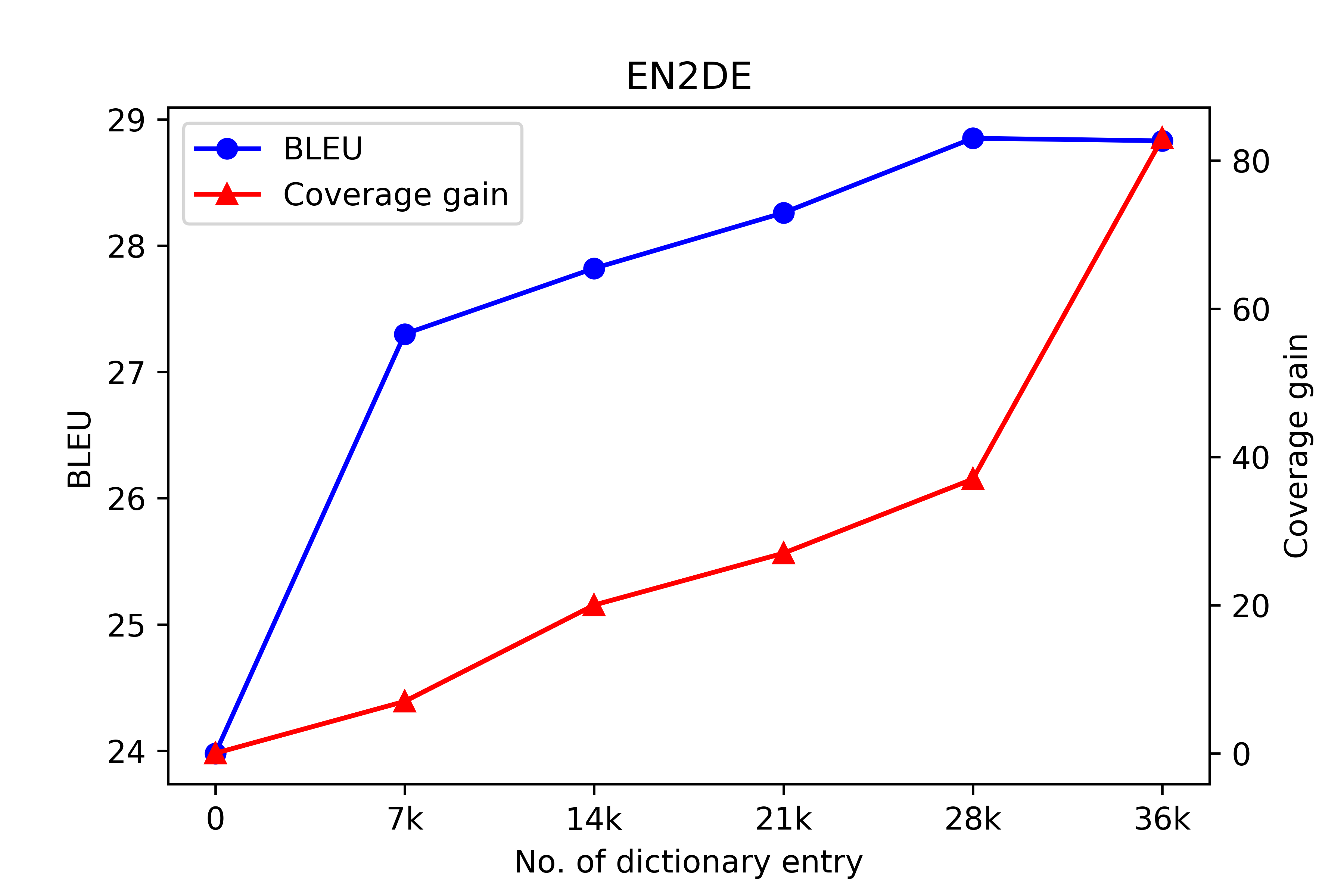}
    \caption{The effects of DDA under various sizes of IND dictionaries for EN2DE models.}
    \label{figure3}
\end{figure}

The results of an analysis of the gain of enhanced domain coverage of the DDA-generated pseudo-IND corpus $ED((DDA-G) \cap Test)$ over that of the OOD training data $ED(Train \cap Test)$ are shown in Figure~\ref{figure2} for EN2FR models. It demonstrates a solid positive correlation between the gain of the $ED$ and BLEU scores of the DDA-enhanced models. More domain-specific terms are identified with the increase of the number of the IND dictionary entry. A similar correlation has been identified for EN2DE models (Figure~\ref{figure3}).

\begin{figure}[h]
    \centering
    \includegraphics[width=0.47\textwidth]{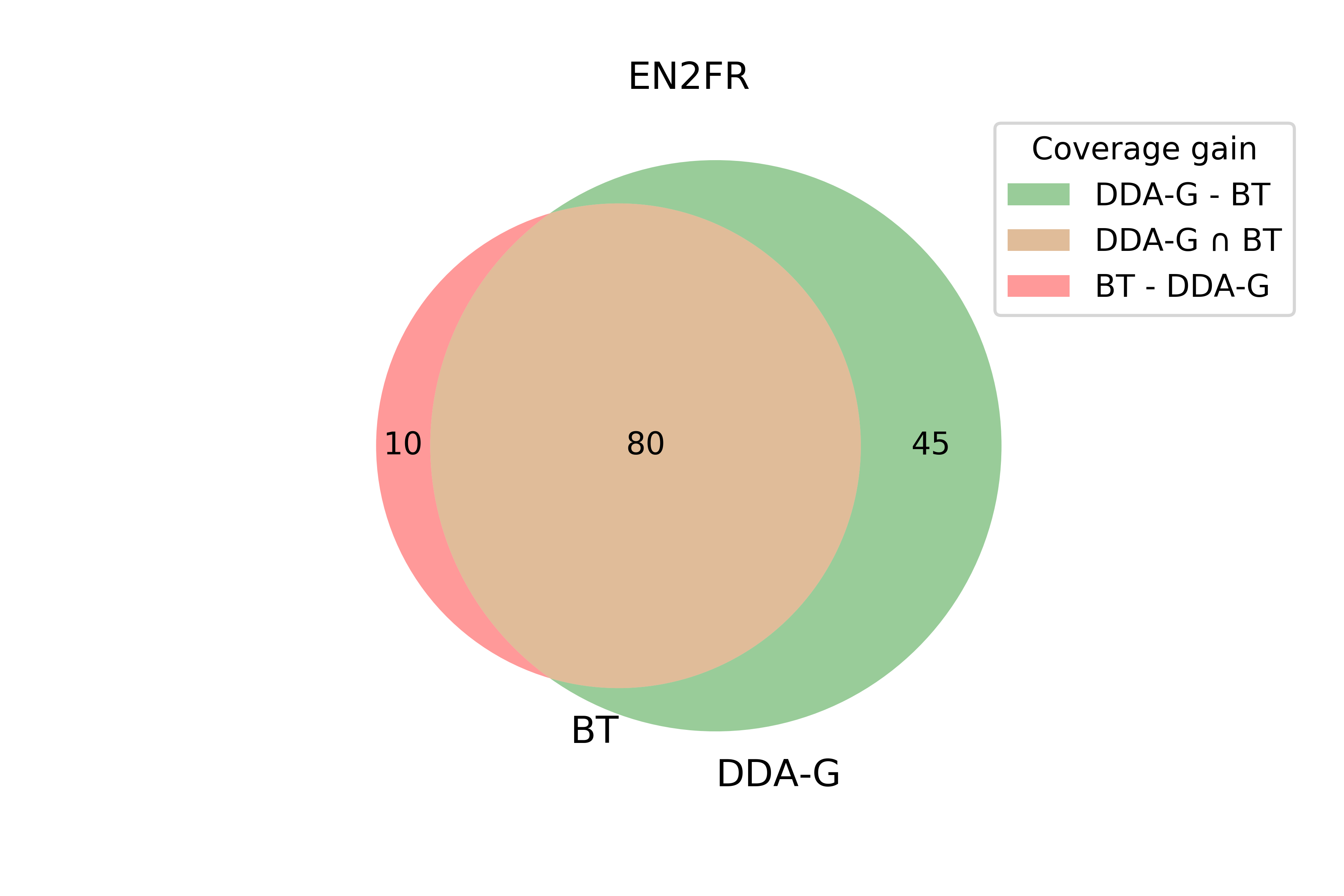}
    \caption{A comparison of the gain of enhanced domain coverage of the DDA-G data with that of a BT-based approach for an EN2FR model. }
    \label{figure4}
\end{figure}

A comparison study is performed for the gain of enhanced domain coverage of DDA-G $(ED((DDA-G) \cap Test))$ and that of the data generated by a BT-based approach $(ED(BT \cap Test))$ over a common base, which is the enhanced domain coverage for the OOD training data $(ED(Train \cap Test))$. It is noted the number of IND terms shown here indicates the extra number of dictionary terms (offered by a BT-based approach and DDA) appears in the test set compared to the terms from the intersect of OOD training data and the test data. DDA-G produces more IND terms (as the green section) than a BT-based approach with their relationships shown in Figure~\ref{figure4}.

\begin{table*}[h]
\centering
\begin{tabular}{lll}
\hline
\multicolumn{3}{c}{\textbf{IND Terms Captured (EN)}} \\ \hline
\textbf{Number of Gram} & \multicolumn{1}{c}{\textbf{BT - DDA-G}} & \multicolumn{1}{c}{\textbf{DDA-G - BT}} \\ \hline
1-gram & \begin{tabular}[c]{@{}l@{}}`SGPT', `aliskiren', \\ `SGOT',`epistaxis', \\ `diaphoresis', \\`premedication'\end{tabular} & \begin{tabular}[c]{@{}l@{}}`valproate', `antiarrhythmics',\\  `mucositis',`Folliculitis', \\ `leiomyoma', `astemizole', \\ `enteropathy', `Atelectasis', \\ `endophthalmitis', `DNAgyrase', \\ ......\end{tabular} \\ \hline
2-gram & \begin{tabular}[c]{@{}l@{}}`cognitive disorder', \\ `Dry mouth', `Muscle pain', \\ `ST segment'\end{tabular} & \begin{tabular}[c]{@{}l@{}}`interstitial nephritis', \\ `Electrolyte imbalance', \\ 'Diphtheria toxoid', \\ 'Steroid myopathy',\\ ......\end{tabular} \\ \hline
3-gram & - & \begin{tabular}[c]{@{}l@{}}`Bone marrow depression', \\ `Renal artery stenosis'\end{tabular} \\ \hline
4-gram & - & `Lower respiratory tract infection' \\ \hline
\end{tabular}
\caption{\label{example2-table} Examples of distinctive IND terms generated by a BT-based approach and DDA.}
\end{table*}

\begin{figure}[h]
    \centering
    \includegraphics[width=0.47\textwidth]{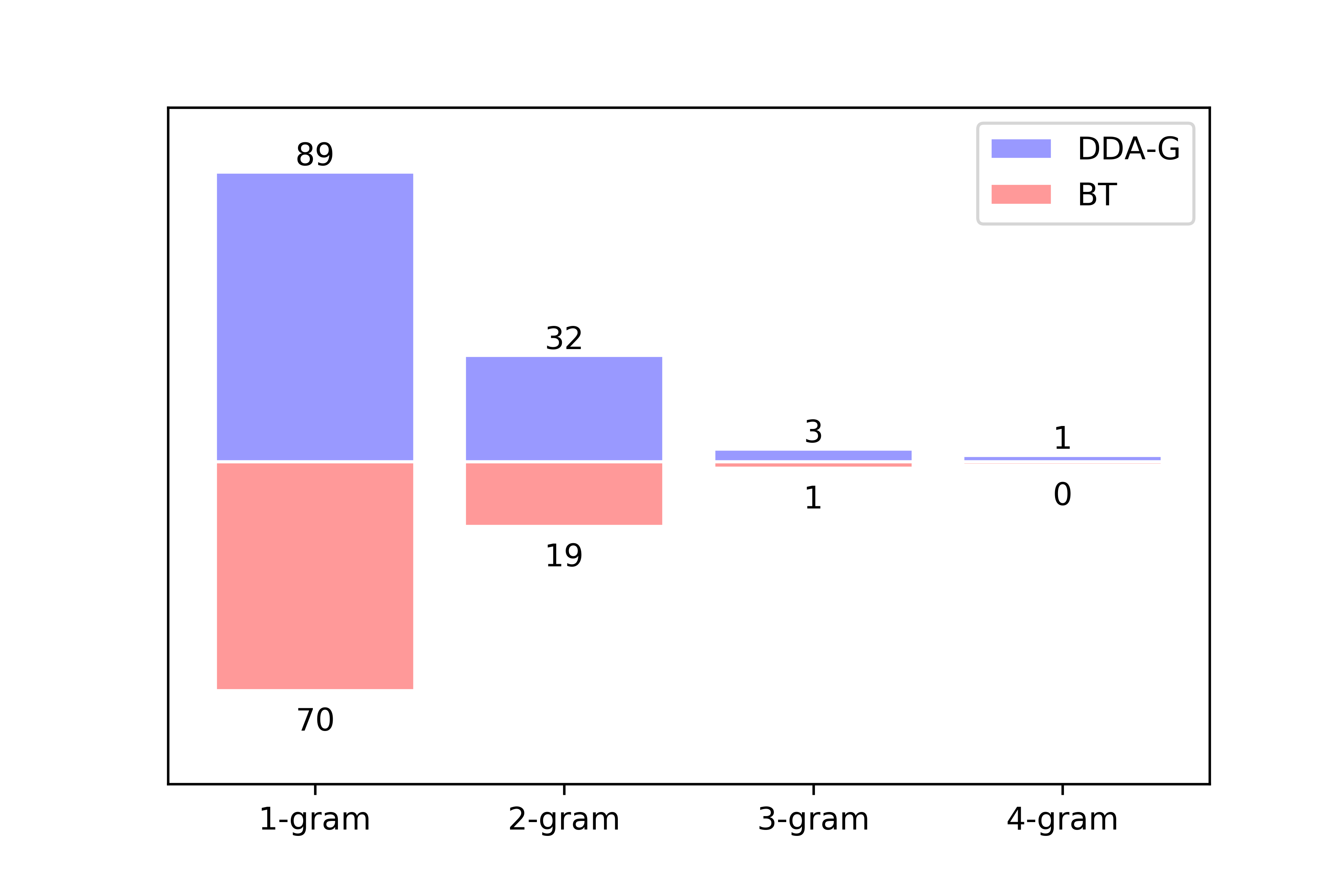}
    \caption{A comparison of 4-gram distributions for IND terms generated by a BT-based approach and DDA-G for an EN2FR model.}
    \label{figure5}
\end{figure}

Among the above IND terms generated by a BT-based method and DDA, respectively, a majority of them consist of unigrams and bigrams phrases. As illustrated in Figure~\ref{figure5}, DDA-G terms lead in all four bins in the distribution. It is an indication of how well DDA performs complementing the BT-based approach. Given that the 45 new IND terms (the green section in Figure~\ref{figure4}) consisting of IND terms across 1-gram (unigram) to 4-gram, we can conclude that DDA provides an effective complementary IND coverage enhancements to a BT-based approach. Results captured in Table~\ref{example2-table} verify this conclusion.

\section{Conclusion}
In this study, we propose a dictionary-based data augmentation (DDA) method for cross-domain NMT. This approach enables a useful and practical way of automatically generating large-scale pseudo-IND parallel corpora from widely available IND dictionaries and OOD bitext. The generated data are used to enhance OOD trained baseline NMT models. Extensive experiments are performed to various NMT models on four translation directions covering English $\leftrightarrow$ French and English $\leftrightarrow$ German language pairs across a general domain and the medical domain.  The results demonstrate consistent significant improvements in BLEU scores over the baseline models. When combined with back-translation, the proposed method can improve the cross-domain translation performance. DDA is confirmed to be a viable complement to BT-based methods in data augmentation for cross-domain NMT. A further analysis unveils that the improvement is associated with the gain of enhanced domain coverage produced by DDA.



\bibliography{DDA-G}

\begin{thebibliography}{27}
\expandafter\ifx\csname natexlab\endcsname\relax\def\natexlab#1{#1}\fi

\bibitem[{Arcan and Buitelaar(2017)}]{arcan}
Mihael Arcan and Paul Buitelaar. 2017.
\newblock \href {http://arxiv.org/abs/1709.02184} {Translating domain-specific
  expressions in knowledge bases with neural machine translation}.
\newblock \emph{CoRR}, abs/1709.02184.

\bibitem[{Cheng et~al.(2016)Cheng, Xu, He, He, Wu, Sun, and
  Liu}]{cheng-etal-2016-semi}
Yong Cheng, Wei Xu, Zhongjun He, Wei He, Hua Wu, Maosong Sun, and Yang Liu.
  2016.
\newblock \href {https://doi.org/10.18653/v1/P16-1185} {Semi-supervised
  learning for neural machine translation}.
\newblock In \emph{Proceedings of the 54th Annual Meeting of the Association
  for Computational Linguistics (Volume 1: Long Papers)}, pages 1965--1974,
  Berlin, Germany. Association for Computational Linguistics.

\bibitem[{Chu and Wang(2018)}]{Chu2018}
Chenhui Chu and Rui Wang. 2018.
\newblock \href {https://www.aclweb.org/anthology/C18-1111} {A survey of domain
  adaptation for neural machine translation}.
\newblock In \emph{Proceedings of the 27th International Conference on
  Computational Linguistics}, pages 1304--1319, Santa Fe, New Mexico, USA.
  Association for Computational Linguistics.

\bibitem[{Currey et~al.(2017)Currey, Barone, and Heafield}]{Currey2017CopiedMD}
Anna Currey, Antonio Valerio~Miceli Barone, and Kenneth Heafield. 2017.
\newblock \href {https://www.aclweb.org/anthology/W17-4715} {Copied monolingual
  data improves low-resource neural machine translation}.
\newblock In \emph{Proceedings of the Conference on Machine Translation (WMT)},
  pages 148--156, Copenhagen, Denmark. Association for Computational
  Linguistics.

\bibitem[{Devlin et~al.(2018)Devlin, Chang, Lee, and
  Toutanova}]{devlin2018bert}
Jacob Devlin, Ming-Wei Chang, Kenton Lee, and Kristina Toutanova. 2018.
\newblock Bert: Pre-training of deep bidirectional transformers for language
  understanding.
\newblock \emph{arXiv preprint arXiv:1810.04805}.

\bibitem[{Domhan and Hieber(2017)}]{domhan}
Tobias Domhan and Felix Hieber. 2017.
\newblock \href {https://doi.org/10.18653/v1/D17-1158} {Using target-side
  monolingual data for neural machine translation through multi-task learning}.
\newblock In \emph{Proceedings of the 2017 Conference on Empirical Methods in
  Natural Language Processing}, pages 1500--1505, Copenhagen, Denmark.
  Association for Computational Linguistics.

\bibitem[{Dyer et~al.(2013)Dyer, Chahuneau, and Smith}]{dyer-etal-2013-simple}
Chris Dyer, Victor Chahuneau, and Noah~A. Smith. 2013.
\newblock \href {https://www.aclweb.org/anthology/N13-1073} {A simple, fast,
  and effective reparameterization of {IBM} model 2}.
\newblock In \emph{Proceedings of the 2013 Conference of the North {A}merican
  Chapter of the Association for Computational Linguistics: Human Language
  Technologies}, pages 644--648, Atlanta, Georgia. Association for
  Computational Linguistics.

\bibitem[{Fadaee et~al.(2017)Fadaee, Bisazza, and
  Monz}]{fadaee-bisazza-monz:2017:Short2}
Marzieh Fadaee, Arianna Bisazza, and Christof Monz. 2017.
\newblock \href {http://aclweb.org/anthology/P17-2090} {Data augmentation for
  low-resource neural machine translation}.
\newblock In \emph{Proceedings of the 55th Annual Meeting of the Association
  for Computational Linguistics (Volume 2: Short Papers)}, pages 567--573,
  Vancouver, Canada. Association for Computational Linguistics.

\bibitem[{Hokamp and Liu(2017)}]{hokamp-liu-2017-lexically}
Chris Hokamp and Qun Liu. 2017.
\newblock \href {https://doi.org/10.18653/v1/P17-1141} {Lexically constrained
  decoding for sequence generation using grid beam search}.
\newblock In \emph{Proceedings of the 55th Annual Meeting of the Association
  for Computational Linguistics (Volume 1: Long Papers)}, pages 1535--1546,
  Vancouver, Canada. Association for Computational Linguistics.

\bibitem[{Hu et~al.(2019)Hu, Xia, Neubig, and Carbonell}]{hu-etal-2019-domain}
Junjie Hu, Mengzhou Xia, Graham Neubig, and Jaime Carbonell. 2019.
\newblock \href {https://www.aclweb.org/anthology/P19-1286} {Domain adaptation
  of neural machine translation by lexicon induction}.
\newblock In \emph{Proceedings of the 57th Annual Meeting of the Association
  for Computational Linguistics}, pages 2989--3001, Florence, Italy.
  Association for Computational Linguistics.

\bibitem[{Johnson et~al.(2017)Johnson, Douze, and J{\'e}gou}]{JDH17}
Jeff Johnson, Matthijs Douze, and Herv{\'e} J{\'e}gou. 2017.
\newblock Billion-scale similarity search with gpus.
\newblock \emph{arXiv preprint arXiv:1702.08734}.

\bibitem[{Kirkpatrick et~al.(2017)Kirkpatrick, Pascanu, Rabinowitz, Veness,
  Desjardins, Rusu, Milan, Quan, Ramalho, Grabska-Barwinska, Hassabis, Clopath,
  Kumaran, and Hadsell}]{Kirkpatrick3521}
James Kirkpatrick, Razvan Pascanu, Neil Rabinowitz, Joel Veness, Guillaume
  Desjardins, Andrei~A. Rusu, Kieran Milan, John Quan, Tiago Ramalho, Agnieszka
  Grabska-Barwinska, Demis Hassabis, Claudia Clopath, Dharshan Kumaran, and
  Raia Hadsell. 2017.
\newblock \href {https://doi.org/10.1073/pnas.1611835114} {Overcoming
  catastrophic forgetting in neural networks}.
\newblock \emph{Proceedings of the National Academy of Sciences},
  114(13):3521--3526.

\bibitem[{Koehn et~al.(2007)Koehn, Hoang, Birch, Callison-Burch, Federico,
  Bertoldi, Cowan, Shen, Moran, Zens, Dyer, Bojar, Constantin, and
  Herbst}]{moses}
Philipp Koehn, Hieu Hoang, Alexandra Birch, Chris Callison-Burch, Marcello
  Federico, Nicola Bertoldi, Brooke Cowan, Wade Shen, Christine Moran, Richard
  Zens, Chris Dyer, Ondrej Bojar, Alexandra Constantin, and Evan Herbst. 2007.
\newblock \href {https://www.aclweb.org/anthology/P07-2045} {{M}oses: Open
  source toolkit for statistical machine translation}.
\newblock In \emph{Proceedings of the 45th Annual Meeting of the Association
  for Computational Linguistics Companion Volume Proceedings of the Demo and
  Poster Sessions}, pages 177--180, Prague, Czech Republic. Association for
  Computational Linguistics.

\bibitem[{Koehn and Knowles(2017)}]{Koehn2017}
Philipp Koehn and Rebecca Knowles. 2017.
\newblock \href {https://www.aclweb.org/anthology/W17-3204} {Six challenges for
  neural machine translation}.
\newblock In \emph{Proceedings of the First Workshop on Neural Machine
  Translation}, pages 28--39, Vancouver, Canada. Association for Computational
  Linguistics.

\bibitem[{Lample et~al.(2018)Lample, Conneau, Ranzato, Denoyer, and
  Jégou}]{lample2018word}
Guillaume Lample, Alexis Conneau, Marc'Aurelio Ranzato, Ludovic Denoyer, and
  Hervé Jégou. 2018.
\newblock \href {https://openreview.net/forum?id=H196sainb} {Word translation
  without parallel data}.
\newblock In \emph{International Conference on Learning Representations}.

\bibitem[{Ott et~al.(2019)Ott, Edunov, Baevski, Fan, Gross, Ng, Grangier, and
  Auli}]{ott2019fairseq}
Myle Ott, Sergey Edunov, Alexei Baevski, Angela Fan, Sam Gross, Nathan Ng,
  David Grangier, and Michael Auli. 2019.
\newblock fairseq: A fast, extensible toolkit for sequence modeling.
\newblock In \emph{Proceedings of NAACL-HLT 2019: Demonstrations}.

\bibitem[{Post(2018)}]{post-2018-call}
Matt Post. 2018.
\newblock \href {https://www.aclweb.org/anthology/W18-6319} {A call for clarity
  in reporting {BLEU} scores}.
\newblock In \emph{Proceedings of the Third Conference on Machine Translation:
  Research Papers}, pages 186--191, Belgium, Brussels. Association for
  Computational Linguistics.

\bibitem[{Sennrich et~al.(2016{\natexlab{a}})Sennrich, Haddow, and
  Birch}]{sennrich2016a}
Rico Sennrich, Barry Haddow, and Alexandra Birch. 2016{\natexlab{a}}.
\newblock \href {https://doi.org/10.18653/v1/P16-1009} {Improving neural
  machine translation models with monolingual data}.
\newblock In \emph{Proceedings of the 54th Annual Meeting of the Association
  for Computational Linguistics}, pages 86--96, Berlin, Germany. Association
  for Computational Linguistics.

\bibitem[{Sennrich et~al.(2016{\natexlab{b}})Sennrich, Haddow, and
  Birch}]{sennrich2016b}
Rico Sennrich, Barry Haddow, and Alexandra Birch. 2016{\natexlab{b}}.
\newblock \href {https://doi.org/10.18653/v1/P16-1162} {Neural machine
  translation of rare words with subword units}.
\newblock In \emph{Proceedings of the 54th Annual Meeting of the Association
  for Computational Linguistics (Volume 1: Long Papers)}, pages 1715--1725,
  Berlin, Germany. Association for Computational Linguistics.

\bibitem[{Thompson et~al.(2019)Thompson, Knowles, Zhang, Khayrallah, Duh, and
  Koehn}]{thompson-etal-2019-hablex}
Brian Thompson, Rebecca Knowles, Xuan Zhang, Huda Khayrallah, Kevin Duh, and
  Philipp Koehn. 2019.
\newblock \href {https://doi.org/10.18653/v1/D19-1142} {{HABL}ex: Human
  annotated bilingual lexicons for experiments in machine translation}.
\newblock In \emph{Proceedings of the 2019 Conference on Empirical Methods in
  Natural Language Processing and the 9th International Joint Conference on
  Natural Language Processing (EMNLP-IJCNLP)}, pages 1382--1387, Hong Kong,
  China. Association for Computational Linguistics.

\bibitem[{Vaswani et~al.(2017)Vaswani, Shazeer, Parmar, Uszkoreit, Jones,
  Gomez, Kaiser, and Polosukhin}]{Vaswani2017}
Ashish Vaswani, Noam Shazeer, Niki Parmar, Jakob Uszkoreit, Llion Jones,
  Aidan~N Gomez, \L~ukasz Kaiser, and Illia Polosukhin. 2017.
\newblock \href
  {http://papers.nips.cc/paper/7181-attention-is-all-you-need.pdf} {Attention
  is all you need}.
\newblock In I.~Guyon, U.~V. Luxburg, S.~Bengio, H.~Wallach, R.~Fergus,
  S.~Vishwanathan, and R.~Garnett, editors, \emph{Advances in Neural
  Information Processing Systems 30}, pages 5998--6008. Curran Associates, Inc.

\bibitem[{Wang et~al.(2017)Wang, Finch, Utiyama, and Sumita}]{wang-etal-2017}
Rui Wang, Andrew Finch, Masao Utiyama, and Eiichiro Sumita. 2017.
\newblock \href {https://doi.org/10.18653/v1/P17-2089} {Sentence embedding for
  neural machine translation domain adaptation}.
\newblock In \emph{Proceedings of the 55th Annual Meeting of the Association
  for Computational Linguistics (Volume 2: Short Papers)}, pages 560--566,
  Vancouver, Canada. Association for Computational Linguistics.

\bibitem[{van~der Wees et~al.(2017)van~der Wees, Bisazza, and
  Monz}]{van-der-wees-etal-2017-dynamic}
Marlies van~der Wees, Arianna Bisazza, and Christof Monz. 2017.
\newblock \href {https://doi.org/10.18653/v1/D17-1147} {Dynamic data selection
  for neural machine translation}.
\newblock In \emph{Proceedings of the 2017 Conference on Empirical Methods in
  Natural Language Processing}, pages 1400--1410, Copenhagen, Denmark.
  Association for Computational Linguistics.

\bibitem[{Xiao(2018)}]{xiao2018bertservice}
Han Xiao. 2018.
\newblock bert-as-service.
\newblock \url{https://github.com/hanxiao/bert-as-service}.

\bibitem[{Xing et~al.(2015)Xing, Wang, Liu, and
  Lin}]{xing-etal-2015-normalized}
Chao Xing, Dong Wang, Chao Liu, and Yiye Lin. 2015.
\newblock \href {https://doi.org/10.3115/v1/N15-1104} {Normalized word
  embedding and orthogonal transform for bilingual word translation}.
\newblock In \emph{Proceedings of the 2015 Conference of the North {A}merican
  Chapter of the Association for Computational Linguistics: Human Language
  Technologies}, pages 1006--1011, Denver, Colorado. Association for
  Computational Linguistics.

\bibitem[{Zhang and Zong(2016{\natexlab{a}})}]{zhangandzong2016:2}
Jiajun Zhang and Chengqing Zong. 2016{\natexlab{a}}.
\newblock \href {http://arxiv.org/abs/1610.07272} {Bridging neural machine
  translation and bilingual dictionaries}.
\newblock \emph{CoRR}, abs/1610.07272.

\bibitem[{Zhang and Zong(2016{\natexlab{b}})}]{zhangandzong2016:1}
Jiajun Zhang and Chengqing Zong. 2016{\natexlab{b}}.
\newblock \href {https://aclweb.org/anthology/D16-1160} {Exploiting source-side
  monolingual data in neural machine translation}.
\newblock In \emph{Proceedings of the 2016 Conference on Empirical Methods in
  Natural Language Processing}, pages 1535--1545, Austin, Texas, USA.
  Association for Computational Linguistics.

\end{thebibliography}
\bibliographystyle{acl_natbib}

\appendix

\section{Appendices}
\label{sec:appendix}
Table~\ref{result2-table} shows that deduplicating EMEA  leads to higher BLEU scores.  

\begin{table}[h]
\centering
\begin{tabular}{lcc}
\hline
\multicolumn{3}{c}{\textbf{Effects of Deduplication for EMEA }} \\ \hline
\begin{tabular}[l]{@{}l@{}} \textbf{Translation}\\ \textbf{Direction} \end{tabular} & \begin{tabular}[l]{@{}c@{}} \textbf{Baseline +} \\ \textbf{EMEA (ori.)}\end{tabular} & \begin{tabular}[l]{@{}c@{}}\textbf{Baseline +} \\ \textbf{EMEA (ded.)}\end{tabular} \\ \hline
\textbf{EN2FR} & 44.46 & 45.90 (\textbf{+1.44}) \\ 
\textbf{FR2EN} & 51.66 & 53.38 (\textbf{+1.72}) \\
\textbf{EN2DE} & 46.13 & 48.71 (\textbf{+2.58}) \\
\textbf{DE2EN} & 53.23 & 55.68 (\textbf{+2.45}) \\
\hline
\end{tabular}
\caption{\label{result2-table} Experimental results for NMT models trained with OOD corpora, fine-tuned with the original EMEA (ori.) and the deduplicated EMEA (ded.). }
\end{table}


\section{Supplemental Material}
\label{sec:supplemental}
Scripts and data are available at https://github.com/nlp-team/DA\_NMT.

\end{document}